\pdfoutput=1

\documentclass[11pt]{article}

\usepackage[]{acl}

\usepackage{times}
\usepackage{latexsym}
\usepackage{graphicx}
\usepackage{enumerate}
\usepackage[T1]{fontenc}

\usepackage[utf8]{inputenc}

\usepackage{microtype}
\usepackage{booktabs}
\usepackage{listings}
\usepackage{tikz}
\usepackage[framemethod=TikZ]{mdframed}
\usepackage{multirow}
\usepackage{algpseudocode}
\usepackage{minted}
\usepackage{algorithm}
\usepackage{csquotes}
\usepackage{colortbl}
\usepackage{rotating}
\usepackage{adjustbox}
\usepackage{amsmath}
\usepackage{amssymb}
\usepackage{soul}
\usepackage{enumitem}
\usepackage{inconsolata}

\newcommand{\ctext}[3][RGB]{%
  \begingroup
  \definecolor{hlcolor}{#1}{#2}\sethlcolor{hlcolor}%
  \hl{#3}%
  \endgroup
}

\title{Face the Facts!\\
Evaluating RAG-based Pipelines for Professional Fact-Checking}

\author{Daniel Russo$^{1,2}$, Stefano Menini$^{1}$, Jacopo Staiano$^{2}$, Marco Guerini$^{1}$ \\
        $^{1}$Fondazione Bruno Kessler, Italy \\
        $^{2}$University of Trento, Italy \\
        \texttt{\{drusso, menini, guerini\}@fbk.eu}, \texttt{jacopo.staiano@unitn.it}
}

\begin{document}
\maketitle
\begin{abstract}
Natural Language Processing and Generation systems have recently shown the potential to complement and streamline the costly and time-consuming job of professional fact-checkers. In this work, we lift several constraints of current state-of-the-art pipelines for automated fact-checking based on the Retrieval-Augmented Generation (RAG) paradigm. Our goal is to benchmark, following professional fact-checking practices,
RAG-based methods for the generation of verdicts - i.e., short texts discussing the veracity of a claim - evaluating them on stylistically complex claims and heterogeneous, yet reliable, knowledge bases. Our findings show a complex landscape, where, for example, LLM-based retrievers outperform other retrieval techniques, though they still struggle with heterogeneous knowledge bases; larger models excel in verdict faithfulness, while smaller models provide better context adherence, with human evaluations favouring zero-shot and one-shot approaches for informativeness, and fine-tuned models for emotional alignment.
\end{abstract}

\begin{figure*}
      \centering
      \includegraphics[width=.9\textwidth]{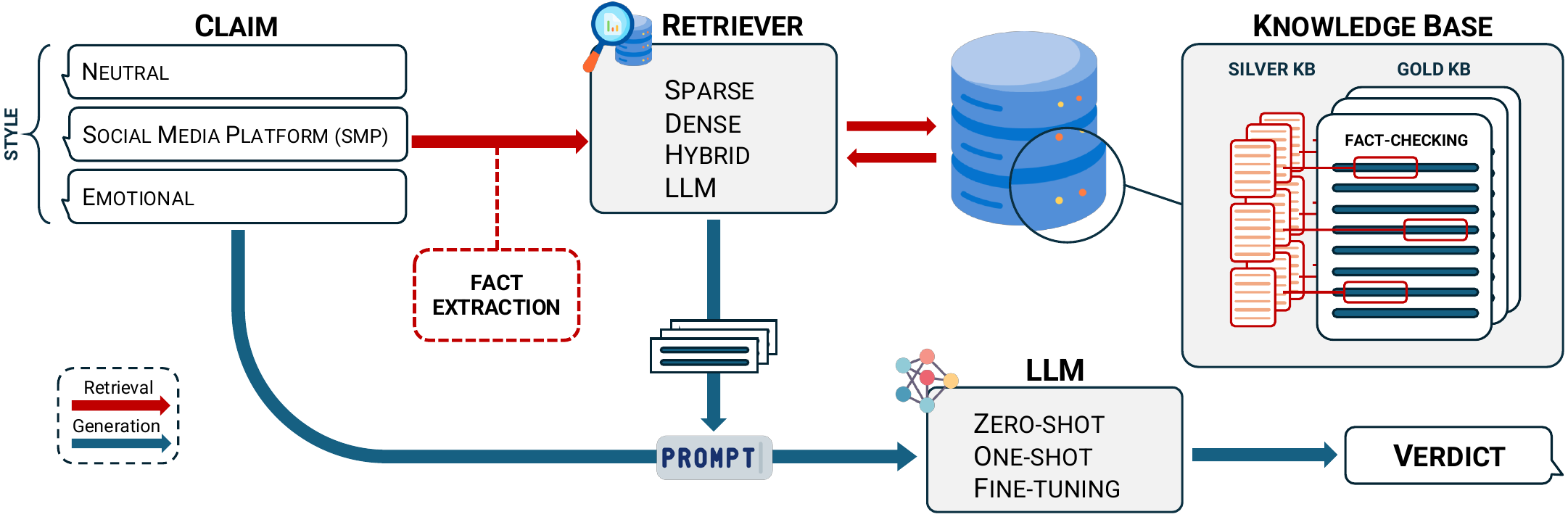}
      \caption{Visual representation of our RAG-based experimental design (the steps for retrieval and generation are indicated by the red and blue lines, respectively). We explored various configurations to tackle increasingly realistic scenarios across different claim styles (neutral, SMP, emotional) and Knowledge Bases (Gold vs. Silver), as well as varying computational demands through multiple retriever architectures (sparse, dense, hybrid, and LLM-based) and five distinct LLMs generation setups (zero-shot, one-shot, fine-tuning).}
      \label{fig:exp_design}
\end{figure*}

\section{Introduction}

Despite the efforts to validate the accuracy of online content, professional fact-checkers are increasingly struggling to keep up with the rapid spread of misinformation \citep{lewis2008quality, adair2017progress, godler2017journalistic, wang2018}. 
Therefore, Natural Language Processing (NLP) has been proposed as a viable solution to partially automate the costly process of verifying misleading claims online \cite{vlachos2014fact}.
Within this context, the task of \textit{verdict production}, i.e., explaining why a claim is true or false, stands as one of the most challenging \citep{kotonya2020explainable,guo2022survey}.

Framing verdict production as a summarization task over fact-checking articles is a suitable solution due to the possibility of generating highly readable verdicts even for non-expert users \citep{atanasova2024generating, kotonya-toni-2020-explainable-health, russo-etal-2023-benchmarking}.
Despite their promising results, summarization-based approaches suffer from two main limitations: \emph{(i)} they rely on the assumption that a fact-checking article always exists for a given claim; and \emph{(ii)}, they further assume that claims are already paired with a fact-checking article, which is typical in fact-checking websites but not on social media platforms, where most of the misinformation spreads \citep{science_of_fn}.

Grounding textual generation on retrieved evidence, an approach named Retrieval-Augmented Generation (RAG), has been shown to be effective for knowledge-intensive tasks like fact verification \citep{lewis-2020-RAG, Liao-2023-MUSER, xu2023retrieval}; it also allows addressing the limitations of previous summarization approaches, e.g., the assumption that the claims are already paired with gold fact-checking articles. Moreover, RAG-based approaches have proven useful in reducing potential factual inconsistencies, often referred to as \textit{``hallucinations''} \citep{zellers2019defending, solaiman2019release}, during text generation \citep{lewis-2020-RAG}, making them attractive for fact-checking tasks. Thus, researchers have increasingly adopted RAG to enhance the accuracy of the generated verdicts \citep{10.1162/tacl_a_00649, yao-etal-2023-end}.

Current studies depend on fact-checking websites, resulting in verdicts characterized by formal and dry language. This style contrasts sharply with the language used on Social Media Platforms (SMPs), which is typically more complex and includes \textit{noise} such as %
personal commentary, or emotional content that surrounds the core fact. 
Such mismatch might pose serious issues when countering misinformation online \citep{Colliander-2019-this}. 
We challenge these common assumptions
on verdict generation, by testing RAG-based pipelines across scenarios that progressively approximate professional fact-checking practices.

To this end, we present a thorough evaluation of verdict production across several dimensions of a RAG pipeline, testing key configurations for each of them (see Figure~\ref{fig:exp_design}): %
\begin{enumerate}[noitemsep, leftmargin=*]
    \item Three claim styles differing in realism: \textsl{neutral} (journalistic), \textsl{SMP} (social media-like), and \textsl{emotional} (SMP enriched with an emotional component) taken from VerMouth dataset \citep{russo-etal-2023-countering};
    \item Four retrieval methods with varying computational costs: \textsl{sparse}, \textsl{dense}, \textsl{hybrid}, and \textsl{LLM-based};
    \item Two claim pre-processing settings, with and without fact extraction (to simplify complex claims written in SMP and emotional style before the retrieval step);
    \item Five LLMs varying in size and training for verdict generation;
    \item Three generation setups: \textsl{zero-shot}, \textsl{one-shot}, and \textsl{fine-tuned};
    \item Two types of knowledge bases: a \textsl{gold KB} (with verified fact-checks) and a \textsl{silver KB} (comprising only the relevant and reliable sources used to write fact-checking articles);
    \item Two document storage strategies: retrieving \textsl{full articles} or smaller \textsl{chunks}.
\end{enumerate}
We show that LLM-based retrievers consistently outperform other methods, though they face challenges with silver knowledge bases. Dense retrievers manage stylistic variations of the claim effectively but fall short compared to LLMs, whereas sparse retrievers exhibit high sensitivity to noise present in emotional and SMP claims. Hybrid approaches and query pre-processing improve performance. 
Turning to generation, larger models excel in faithfulness and alignment with gold verdicts, while smaller ones are more consistent in context adherence. Fine-tuning boosts similarity of the generated verdict to the gold but reduces contextual accuracy, with human evaluations favouring verdicts generated under zero/one-shot strategies for informativeness and fine-tuning for emotional alignment.\footnote{Code and data are publicly available on GitHub: \href{https://github.com/drusso98/face-the-facts}{https://github.com/drusso98/face-the-facts}.}

\section{Related Work}

Early approaches to verdict generation leveraged either attention modules to highlight salient tokens from the evidence text \citep{popat2018declare,shu-etal-2019-dEFEND, lu-li-2020-gcan, wu-etal-2020-dtca,yang-etal-2019-XFake}, or Horn Rules to reason upon structured knowledge bases \citep{Gad-Elrab-etal-2019-ExFact, ahmadi2019explainable}. However, both lack readability, being hard to interpret by common users \citep{guo2022survey}.
To overcome this issue, researchers started casting verdict production as a summarization task over fact-checking articles, either through extractive \citep{atanasova-etal-2020-generating-fact}, abstractive \citep{kotonya2020explainable,Stammbach2020eFEVEREA}, or hybrid \citep{russo-etal-2023-benchmarking} summarization pipelines. More recently, \citet{he-etal-2023-reinforcement} introduced a reinforcement learning-based framework for generating counter-misinformation responses to social media content.

Ad-hoc data collection strategies for verdict generation rely either on synthetic data generation, like e-FEVER \citep{Stammbach2020eFEVEREA}, or on %
journalistic sources, such as LIARPLUS \citep{alhindi-etal-2018-evidence}, PUBHEALTH \citep{kotonya-toni-2020-explainable-health}, RU22Fact \citep{zeng-etal-2024-ru22fact}, LIAR++, and FullFact \citep{russo-etal-2023-benchmarking}. \citet{russo-etal-2025-euroverdict} proposed EuroVerdict, a multilingual, manually curated dataset for verdict generation. For more realistic, SMP-style claims, \citet{he-etal-2023-reinforcement} developed MisinfoCorrect, and \citet{russo-etal-2023-countering} extended FullFact with VerMouth, incorporating emotional claims and verdicts grounded in trustworthy fact-checking articles.

While the latest approaches provide readable verdicts, they may lack faithfulness due to language models generating factual inaccuracies. Additionally, summarization approaches assume that trustworthy evidence is always available for the claim under inspection. Therefore, RAG-based approaches have been employed to guide the generation of reliable verdicts upon trustworthy evidence previously retrieved from a knowledge base (KB). To this end, \citet{10.1162/tacl_a_00649} proposed JustiLM, a few-shot RAG-based approach for the generation of verdicts for real-world claims, by leveraging both fact-checking articles and auxiliary evidence during model training. \citet{yao-etal-2023-end} developed an end-to-end RAG-based system to perform verdict prediction and production in a multimodal setting jointly. Nevertheless, both studies concentrate on journalistic data and writing styles, without considering the communication style employed on SMP.

Building on the work of \citet{10.1162/tacl_a_00649}, we present an extensive evaluation of RAG pipelines for verdict generation, testing various combinations of retrieval and generation strategies. Progressing toward increasingly realistic scenarios, with respect to the workflow adopted by professional fact-checkers, we address the challenge of handling claims and sources that vary in style and complexity, aiming to closely mimic the kinds of claims that everyday users might encounter on social media platforms. Specifically, we assess the impact of claim writing style on retriever performance, highlighting where differences arise as the style shifts toward that used by SMP users. Finally, we explore the extreme scenario where no fact-checking article exists, relying solely on reliable supporting evidence.

\section{Experimental Design}

In this section, we provide details on the experimental design: from the datasets used, through the retrieval methods adopted, to the configurations of the LLMs employed for verdict generation. 

\subsection{Dataset}

To study the impact of different styles on a RAG-based verdict production task, we used FullFact \citep{russo-etal-2023-benchmarking} and VerMouth \citep{russo-etal-2023-countering} datasets. The two datasets comprise eight different versions of the same claims and verdicts. FullFact provides data written in a journalistic style scraped from \href{www.fullfact.org}{fullfact.org}
while VerMouth proposes the same data rewritten in an SMP style, and also enriched with the six emotional components defined by \citet{Ekman1992AnAF}. 

In both datasets, each claim-verdict pair is linked to a human-written fact-checking article, thus compounding to 8 different versions of the same claim: journalistic style (\textit{neutral} hereafter), \textit{SMP} style, anger, surprise, disgust, joy, fear, and sadness. In VerMouth, verdicts were also rewritten to reflect the various styles and emotions present in the claims. Throughout the paper, we will refer to emotion-styled subsets as \textit{emotional} data.\footnote{More details on the datasets are provided in Appendix \ref{app:dataset_details}.}

\subsection{Retrieval Module}
This comprises three elements: a \textit{query} (a claim in our case), a \textit{knowledge base} (KB), and a \textit{retriever}.

\paragraph{Claim} %
We used claims from FullFact and VerMouth datasets as queries.
The two datasets offer three aligned variations of a claim: neutral, SMP, and emotional.
Due to \textit{noise} in SMP and emotional data, i.e., irrelevant information surrounding the main facts, directly using claims as queries can negatively impact the retrievers' performance.
\textit{Query rewriting}, which transforms context-dependent user queries into self-contained ones, has proven to be an effective approach for enhancing retriever performance \citep{elgohary-etal-2019-unpack, ye-etal-2023-enhancing}. For this reason, we implemented a \textbf{fact extraction module} to simplify claims and remove noise around the main fact we need to retrieve evidence for. In particular, we employed \texttt{Llama-2-13b-chat-hf} in a one-shot learning setup to extract the main facts from all SMP and emotional claims. A manual evaluation of the model's output confirmed the effectiveness of the methodology.\footnote{The full instruction prompt and the manual evaluation of the fact extraction module are reported in Appendix~\ref{app:fact_extraction}.}
An example of an emotional claim and its related fact is provided below.

\begin{quote}
\textit{Unbelievable! Just heard that 53 people have lost their lives in Gibraltar within 10 days of receiving Pfizer's Covid-19 vaccine. This is beyond alarming and I am absolutely furious. How can we trust these vaccines when they're causing more harm than good?! \#PfizerVaccine \#COVID19}
\end{quote}
\begin{quote}
\textit{53 people have lost their lives in Gibraltar within 10 days of receiving Pfizer's Covid-19 vaccine.}
\end{quote}

\paragraph{Retriever} We evaluated several retrieval strategies, with varying computational demands to accommodate the potential computational constraints of the target users: \emph{(i)} sparse: \texttt{BM25} and \texttt{BM25+} \cite{robertson1995okapi}, a popular and effective extension of tf-idf; \emph{(ii)} dense: \texttt{Dragon+} \cite{lin-etal-2023-train} and \texttt{Contriever} \cite{izacard2021unsupervised};
\emph{(iii)} hybrid,
combining \texttt{BM25+} and \texttt{Dragon+} retrievers, using \texttt{BAAI/bge-reranker-large} as a reranker \cite{bge_embedding};
and, \emph{(iv)} an instruction-tuned LLM for text embedding, \texttt{e5-mistral-7b-instruct} \cite{wang2023improving}\footnote{\href{https://hf.co/intfloat/e5-mistral-7b-instruct}{https://hf.co//intfloat/e5-mistral-7b-instruct}}, LLM-Retriever hereafter.

\paragraph{Knowledge Base}
\label{par:knowledge_base}
To build the KB, we employed articles from the FullFact dataset, aligned with VerMouth data. We named this KB as \textbf{Gold KB}.
We experimented with two approaches: \emph{(i)} indexing entire articles (Gold\_KB$_{art}$); \emph{(ii)} %
indexing small portions of each article as separate documents, i.e. \emph{chunks}\footnote{We used the \texttt{LlamaIndex} \cite{Liu_LlamaIndex_2022} sentence splitter, which minimises text fragmentation by keeping sentence integrity, with a maximum chunk token size of 100.} (Gold\_KB$_{chunks}$).

In a realistic scenario, an up-to-date KB of fact-checking articles may not be available, or a fact-checking article might not exist (yet) for a given claim. To approximate this scenario, we leveraged knowledge from reliable sources to build a \textbf{Silver KB}. Specifically, we discarded gold fact-checking articles and extracted the evidence used to write and fact-check claims from FullFact's articles. 
This design choice is grounded in direct collaboration with approximately 20 professional fact-checkers. They emphasized that they do not rely on open web search, but instead consult curated and trustworthy sources, such as the \textit{Google Fact Check Tools} or predefined lists of reputable websites.
The Silver KB thus serves as a faithful proxy for this professional workflow, making our experimental setup more realistic and practically grounded.
The Silver KB was collected by following the URLs present in the articles and getting their textual content.
FullFact articles also typically link to the sources of the claims.
However, the reliability of these sources is questionable; thus, we filtered them out. Also, we ignored all links to social networks (Twitter, Facebook, Instagram, TikTok, and Reddit). Finally, from the remaining URLs,  we extracted the text using the \texttt{Newspaper3k}\footnote{\href{https://github.com/codelucas/newspaper/}{https://github.com/codelucas/newspaper/}} Python library. Statistics related to the extra evidence collection are presented in Appendix~\ref{app:extra_evidence}.

\subsection{Verdict Generation}
\label{subsection:verdict_generation}
For the generation of the verdicts, we tested five LLMs, selected based on differences in sizes or the presence of guardrails: \texttt{Mistral}, in its v1.0 and v2.0 versions \cite{jiang2023mistral}; \texttt{Llama-2} \cite{touvron2023llama}, in its 7B and 13B chat versions;\footnote{\href{https://hf.co/meta-llama/Llama-2-7b-chat-hf}{https://hf.co/meta-llama/Llama-2-7b-chat-hf}; \href{https://hf.co/meta-llama/Llama-2-13b-chat-hf}{https://hf.co/meta-llama/Llama-2-13b-chat-hf}} and \texttt{Llama-3-8B-Instruct}.\footnote{\href{https://hf.co/meta-llama/Meta-Llama-3-8B-Instruct}{https://hf.co/meta-llama/Meta-Llama-3-8B-Instruct}} We combined the claim and the retrieved evidence to prompt the LLM (see Appendix~\ref{app:model_instruction}), and tested generation under different setups, namely
zero-shot, one-shot,
and fine-tuning. For fine-tuning, we employed \texttt{Llama-2-13b}, the best-performing model in zero-shot and one-shot settings.

\begin{table*}[h!]
\small
\centering
\begin{tabular}{@{}llcccccccc@{}}
\toprule
                & & \textbf{sparse} & \textbf{dense} & \textbf{hybrid} & \multicolumn{1}{c|}{\textbf{LLM-Retriever}} & \textbf{sparse} & \textbf{dense} & \textbf{hybrid} & \textbf{LLM-Retriever} \\ \midrule 
                & & \multicolumn{4}{c|}{\cellcolor[HTML]{D9D9D9} \textbf{hit\_rate@1}}         & \multicolumn{4}{c}{\cellcolor[HTML]{D9D9D9} \textbf{mrr@10}}    \\
\multirow{6}{*}{\begin{turn}{90}\textbf{Articles}\end{turn}} & Neutral        & 0.903 & 0.905 & \textbf{0.966} & \multicolumn{1}{c|}{\underline{0.960}} & 0.931    & 0.938    & \underline{0.972}    & \textbf{0.978}   \\
& SMP             & 0.770 & \underline{0.799} & \textbf{0.937} & \multicolumn{1}{c|}{\textbf{0.937}} & 0.817    & 0.866    & \textbf{0.963}    & \underline{0.962}   \\
& Emotional       & 0.778 & 0.839 & \underline{0.905} & \multicolumn{1}{c|}{\textbf{0.938}} & 0.838    & 0.866    & \underline{0.933}    & \textbf{0.964}   \\ \cmidrule(l){2-10} 
& SMP$_{facts}$       & 0.778 & 0.801 & \textbf{0.937} & \multicolumn{1}{c|}{\underline{0.914}} & 0.837    & 0.891    & \textbf{0.963}    & \underline{0.947}   \\
& Emotional$_{facts}$ & 0.835 & 0.846 & \underline{0.905} & \multicolumn{1}{c|}{\textbf{0.932}} & 0.883    & 0.897    & \underline{0.933}    & \textbf{0.958}   \\ \midrule
                & & \multicolumn{4}{c|}{\cellcolor[HTML]{D9D9D9} \textbf{hit\_rate@10}}        & \multicolumn{4}{c}{\cellcolor[HTML]{D9D9D9} \textbf{map@10}} \\
\multirow{6}{*}{\begin{turn}{90}\textbf{Chunks}\end{turn}} & Neutral        & 0.963 & \underline{0.992} & \textbf{1.000} & \multicolumn{1}{c|}{\textbf{1.000}} & 0.392   & 0.552   & \underline{0.573}   & \textbf{0.655}   \\
& SMP             & 0.856 & 0.974 & \underline{0.994} & \multicolumn{1}{c|}{\textbf{1.000}} & 0.275   & 0.484   & \underline{0.536}   & \textbf{0.619}   \\
& Emotional       & 0.904 & 0.977 & \underline{0.994} & \multicolumn{1}{c|}{\textbf{1.000}} & 0.273   & 0.482   & \underline{0.545}   &\textbf{ 0.599}   \\ \cmidrule(l){2-10} 
& SMP$_{facts}$       & 0.905 & 0.972 & \underline{0.994} & \multicolumn{1}{c|}{\textbf{1.000}} & 0.304   & 0.505   & \underline{0.526}   & \textbf{0.601}   \\
& Emotional$_{facts}$ & 0.939 & 0.978 & \underline{0.994} & \multicolumn{1}{c|}{\textbf{0.999}}& 0.345   & 0.518   & \underline{0.552}   & \textbf{0.615}   \\ \bottomrule
\end{tabular}
\caption{Results for the retrieval experiments. We report hit\_rate, mrr, and map for retrieval over the Gold\_KB$_{art}$ (Articles) and the Gold\_KB$_{chunks}$ (Chunks) KBs. SMP$_{facts}$ and Emotional$_{facts}$ indicate input preprocessing with the fact extraction module. The first and second best results for claim style are in bold and underlined, respectively.%
}
\label{tab:retrieval_exp_results}
\end{table*}

\section{Retrieval Experiments} %

\paragraph{Retrieval from Gold KB} We tested the retrievers on FullFact and VerMouth test sets with an increasing number of retrieved documents ($k = 1,\dots, 10$).
For each claim used as a query, we considered as relevant documents the fact-checking article, or its chunks, linked to the claim. 
For space reasons, results on the emotional datasets will be presented in aggregated form, referred to as the \textit{`emotional'} set.
Experiments were carried out integrating into the LlamaIndex \citep{Liu_LlamaIndex_2022} framework either \texttt{Rank-BM25} \citep{rank_bm25} or HuggingFace's models \citep{wolf2020huggingfaces}; retrieval performance was assessed with \texttt{ranx} \cite{ranx}. 

Table \ref{tab:retrieval_exp_results} presents retrieval results for each retrieval approach (sparse, dense, hybrid, LLM-Retriever) across all claim's styles (neutral, SMP, emotional) and fact-extraction pre-processings (SMP$_{facts}$, emotional$_{facts}$) using both KB configurations ($Gold\_KB_{art}$ and $Gold\_KB_{chunks}$). For $Gold\_KB_{art}$, we report hit\_rate@1 and MRR@1 (Mean Reciprocal Rank), as each claim had only one gold related article. For $Gold\_KB_{chunks}$, we report hit\_rate@10 and map@10 (Mean Average Precision) to assess whether the retrievers could consistently include at least one gold chunk among the top 10 and the precision of the retrieval system across different recall levels.\footnote{More details in Appendix~\ref{app:retrieval_results}.} 

For article retrieval, all four retrievers achieved high accuracy on neutral claims, with a hit\_rate@1 above 90\%. When the correct article was not immediately retrieved, they still ranked it highly, as shown by strong MRR@10 scores. 
Performance declined with noisier claims, especially for sparse retrievers, while dense models and the LLM-Retriever showed greater robustness. Hybrid retrieval (combining sparse and dense) performed comparably to the LLM-Retriever.

In chunk retrieval, the LLM-Retriever excelled, achieving a hit\_rate@10 that always included a gold chunk and reaching an average MAP@10 of ~70\%. Sparse retrievers showed low map scores ($\sim$40\% for neutral claims), particularly for SMP and emotional claims ($<$30\%), while dense and hybrid approaches followed trends similar to article retrieval. Thus, the low MAP scores indicate that sparse retrievers must retrieve more chunks from the knowledge base to select the relevant content. However, this comes at the cost of also retrieving more non-relevant content, which could potentially compromise the subsequent generation phase.

Overall, the LLM-Retriever consistently outperforms other approaches. Notably, it remains stable even when exposed to different input claim styles, with minimal degradation when noise is introduced. A paired t-test confirms that the performance gains of LLM-Retrieval are statistically significant compared to other methods, with the exception of the hybrid retriever over  hit\_rate@1. 
Still, it maintains a slightly higher mean score (0.934 vs. 0.925) and lower variance (0.061 vs. 0.069). Dense retrievers perform worse but show robustness to stylistic variations. In contrast, sparse retrievers are significantly affected by data noise, resulting in performance drops across all three datasets: 
we find that the fact extraction module we included yields consistent performance improvements, and particularly helps when using sparse and dense retrievers. 
     
 \paragraph{Retrieval from Silver KB}
\label{par:retr_extra_ev}    
We tested the optimal retriever methodologies from the previous experiments, specifically the LLM-Retriever and the hybrid retriever. As outlined in Section~\ref{par:knowledge_base}, the Silver KB consists of reliable sources that have been extracted from the initial fact-checking articles. The evidence consisted of 9983 chunks, each corresponding to a fact-checking article considered a gold standard during evaluation. 

The results (Table~\ref{fig:extra_ev_results}) show that modifying the knowledge base strongly impacted retrieval performance both across the three datasets and the two retrieval strategies. In particular, the two retrievers exhibited comparable performance, mirroring the behaviour observed in earlier experiments with the Gold KB.
Unlike the previous setting with the Gold KB, the LLM-Retrieval's performance in this context is markedly influenced by the stylistic nature of the claims: neutral formulations consistently yielded higher results, and performance was impacted by the claims' complexity.  
    
Prepending a fact extraction module generally improves retrieval results: even robust retrievers can benefit from preprocessing when dealing with heterogeneous KB (that do not contain gold fact-checking articles) and complex (e.g., emotional) claims.

\begin{table}
\centering
\begin{adjustbox}{max width=\columnwidth}
\begin{tabular}{lccccc}
\toprule
 & \textbf{Neutral} & \textbf{SMP} & \textbf{Emotional} & \textbf{SMP$_{facts}$} & \textbf{Emotional$_{facts}$} \\
\midrule
 LLM-Retriever & 0.683 & 0.652 & 0.637 & 0.689 & 0.671 \\
 hybrid        & 0.683 & 0.652 & 0.631 & 0.602 & 0.652 \\
\bottomrule
\end{tabular}
\end{adjustbox}
\caption{Hit\_rate@10 scores for retrieval with LLM-Retrieval and hybrid retriever over the Silver KB.}
\label{fig:extra_ev_results}
\end{table}

\section{Generation Experiments}

\begin{table*}[t!]
    \centering
    \small
    \resizebox{\textwidth}{!}{%
    \begin{tikzpicture}
    \node (table) [inner sep=0pt] {
    \begin{tabular}{p{3.5cm}p{6.5cm}p{5cm}}
        \addlinespace[0.3em]
          \multicolumn{1}{c}{\textbf{Neutral}}    & \multicolumn{1}{c}{\textbf{SMP-style}}     & \multicolumn{1}{c}{\textbf{Emotional style}}      \\ \midrule
        Face masks can cause low oxygen levels in the body.
        & \#StaySafe alert: Face coverings can lead to decreased O2 in the body, so watch for any signs of trouble breathing. Wear masks only if necessary \& keep an eye out for signs of low oxygen. \#Masks \#OxygenLevels \#Health
        &  I'm really scared about the thought of wearing a face mask causing low oxygen levels in my body. I know it's important to protect others, but I don't want to harm myself in the process. \#COVID19 \#facemasks \#safetyconcerns \\ \addlinespace[0.3em]
        \hline & \\[-1.5ex]
        No, the evidence suggests that masks do not cause low oxygen levels.
        & Thank you for sharing this information. It's important to note that the World Health Organization (WHO) and other medical professionals have consistently stated that face masks are a safe and effective way to reduce the transmission of COVID-19. While it's possible that masks may cause some discomfort or difficulty breathing, there is no evidence to suggest that they can cause a significant decrease in oxygen levels. If you have any concerns about your health or the health of others, it's always best to consult with a medical professional.  \#StaySafe \#FaceMasks \#COVID19
        & I understand your concern about wearing face masks and the potential for low oxygen levels. However, there is no evidence to suggest that face masks cause low oxygen levels. In fact, many fact-checkers have debunked this claim. It's important to prioritize your safety while protecting others, and face masks are a crucial tool in doing so. \#COVID19 \#facemasks \#safety \\ \addlinespace[0.3em]
    \end{tabular}
    };
    \draw [rounded corners=.5em, very thick] (table.north west) rectangle (table.south east);
    \end{tikzpicture}}
    \caption{Example of generations using claims form FullFact and VerMourt (first line), e5-mistral as a retriever, and \texttt{LLaMA-2-13b} fine-tuned for the generation of the verdict (second line).}
    \label{tab:generation_examples_main}
\end{table*}

For verdict generation, the claim and its corresponding retrieved evidence were combined into a prompt fed to the five different LLMs (Section~\ref{subsection:verdict_generation}).
For evidence retrieval, we employed the best-performing retriever (i.e., LLM-Retriever). We tested the five LLMs under three setups: zero-shot, one-shot, and fine-tuned.\footnote{The instructions used are provided in Appendix~\ref{app:model_instruction}.} 
For fine-tuning, we employed the best-performing model in both the zero-shot and one-shot configurations, \texttt{Llama-2-13b}. 

When using the Gold\_KB$_{art}$, we included the top-1 article (i.e., the most relevant) in the prompt, a choice justified by the LLM retriever's remarkable hit\_rate@1 results. Conversely, with retrieval from Gold\_KB$_{chunks}$, we fed the model with 10 retrieved chunks, based on its map@10 performance (see Table~\ref{tab:retrieval_exp_results} and Figure \ref{app:retrieval_results}).\footnote{Examples of generations are provided in Appendix~\ref{app:examples}.}

\paragraph{Automatic Metrics} Inspired by previous works \citep{russo-etal-2023-countering, 10.1162/tacl_a_00649}, we use ROUGE-LSum \cite{lin-2004-rouge}, BARTScore \citep{NEURIPS2021_e4d2b6e6-bartscore}, and SummaC \citep{laban-etal-2022-summac} to evaluate lexical adherence and faithfulness of the generated text to the context provided to the LLM. Further, we used BARTScore, which is unaffected by differences in text length, to compute the semantic similarity (\textit{GoldSim}) between the generated and gold verdicts from FullFact and VerMouth. 
For average performances per dataset and per model, see Tables~\ref{tab:gen_res_average_df} and~\ref{tab:average_llm_chunks}, respectively.\footnote{Complete results can be found in Appendix~\ref{app:generation_results}, Table~\ref{tab:complete_gen_results}.}

\paragraph{Zero-shot and One-shot}
Generations with neutral claims yielded better results compared to the SMP and emotional data in both zero-shot and one-shot experiments, indicating that the complexity of claims affects not only the retrieval phase but also the generation phase. Interestingly, when comparing the similarity between the generated and the gold verdicts, the generations with emotional data produced the best results (Table~\ref{tab:gen_res_average_df}). 
Upon manual inspection, we found %
recurrent patterns in the SMP and emotional data, such as expressions of empathy and politeness %
typical of ChatGPT that was used to generate the manually curated verdicts of VerMouth (``I understand your frustration'', ``It is important to note that''). These patterns were also replicated by the models used in this study. Notably, zero-shot experiments generally outperformed one-shot experiments when results were averaged across all the data (Table~\ref{tab:gen_res_average_df}) and LLMs (Table~\ref{tab:average_llm_chunks}). 
Turning to individual model performances (Table~\ref{tab:average_llm_chunks}), larger models (\texttt{Llama-2-13b}) demonstrated higher faithfulness to the context and similarity to gold verdicts, whereas smaller ones (\texttt{mistral-7b-v0.1}, \texttt{llama3-8b}) showed better contextual adherence in terms of overlaps (ROUGE-LSum) and consistency (SummaC). Misalignments between SummaC and GoldSim were observed, often stemming from the fact that fact-checking articles might contain multiple supporting arguments. When the retriever or LLM selects only a subset, the generated verdict may diverge from the gold verdict in argumentation while remaining contextually accurate.

To sum up, generations with neutral claims outperformed those with emotional data in both zero-shot and one-shot experiments but produced more accurate results when paired with emotional data. In terms of faithfulness to context and similarity to the gold, larger models generally performed better. However, smaller models exhibited superior contextual adherence.

\paragraph{LLM Fine-Tuning}

We fine-tuned \texttt{Llama-2-13b}, the best-performing model in the previous in-context learning experiments, disjointly on the three claim styles. To this end, the model was fed with claims and gold verdicts from the FullFact and VerMouth datasets, complemented by positive and negative contextual information.\footnote{Details are provided in Appendix~\ref{app:training_set_creation}.} We randomly sampled 200 entries from both neutral and SMP training datasets; similarly, to obtain a comparably sized dataset for emotional data, we randomly sampled 35 examples for each of the 6 emotional dimensions available.\footnote{Fine-tuning details are reported in Appendix~\ref{app:fine_tuning_details}.} Examples of generation are reported in Table~\ref{tab:generation_examples_main}.
Fine-tuning results (Tables~\ref{tab:gen_res_average_df} and~\ref{tab:average_llm_chunks}) show that \texttt{LLaMA-2-13b} models improve in faithfulness and similarity w.r.t.  gold verdicts across all datasets and for both full articles and text chunks. However, this comes at the cost of lower ROUGE-L scores: without fine-tuning, the models tend to extract and replicate not only the necessary information but also the exact wording from the context. Thus, fine-tuned models abstract better from the context, as expected, and also show better performance in selecting relevant and reliable information.
Further, after fine-tuning, the emotional models achieved higher similarity scores to the original claims. Akin to the zero and one-shot configurations, inspection of the generated verdicts reveals that these models produce empathetic expressions similar to those found in VerMouth. 

\begin{table*}[h!]
    \centering
    \small
    \begin{adjustbox}{max width=\textwidth}
    \begin{tabular}{@{}clcccc|cccc@{}}
    \toprule
                                        &          & \multicolumn{4}{c|}{\textbf{Articles}}            & \multicolumn{4}{c}{\textbf{Chunks}}               \\ \midrule
                                        &          & \textbf{ROUGE-LSum} & \textbf{BARTScore} & \textbf{SummaC} & \textbf{GoldSim} & \textbf{ROUGE-LSum} & \textbf{BARTScore} & \textbf{SummaC} & \textbf{GoldSim} \\
    \multirow{3}{*}{ \textbf{\shortstack{zero\\shot}}}    & Neutral & \textbf{0.16}       & \textbf{-2.13}    & \textbf{0.35}  & -3.03         & \textbf{0.25}     & \textbf{-2.61}    & \textbf{0.25}  & -3.06           \\
                                        & SMP       & 0.15               & -2.31              & 0.32          & -3.01          & 0.24          & -2.69           & 0.24          & -3.03           \\
                                        & emotional  & 0.14               & -2.48              & 0.31          & \textbf{-2.94} & 0.22          & -2.79           & 0.23          & \textbf{-2.98}  \\ \midrule 
    \multirow{3}{*}{\textbf{\shortstack{one\\shot}}}    & Neutral  & \textbf{0.16}      & \textbf{-2.13}     & \textbf{0.33} & -3.03          & \textbf{0.26} & \textbf{-2.53}  & 0.24          & -2.99           \\
                                        & SMP       & 0.15               & -2.42              & 0.32          & -3.01          & 0.24          & -2.73           & \textbf{0.23} & -2.95           \\
                                        & emotional  & 0.14               & -2.60              & 0.32          & \textbf{-2.95} & 0.22          & -2.85           & \textbf{0.23} & \textbf{-2.91}   \\ \midrule
    \multirow{3}{*}{\textbf{\shortstack{fine\\tuning}}}  & Neutral  & \textbf{0.10}      & \textbf{-1.45}     & \textbf{0.53} & -2.71          & 0.08          &  \textbf{-1.42} & \textbf{0.52} & -2.75            \\
                                        & SMP       & \textbf{0.10}      & -2.30              & 0.33          & -2.58          & \textbf{0.10} & -2.45           & 0.32          & \textbf{-2.63}   \\
                                        & emotional & \textbf{0.10}      & -2.43              & 0.31          & \textbf{-2.48} & \textbf{0.10} & -2.76           & 0.31          & -2.68            \\ \bottomrule 
    \end{tabular}
    \end{adjustbox}
    \caption{Generation results per dataset, \textit{averaged across the LLMs}. Retrieved articles or chunks were employed in the generation. The best results for each generation configuration are in bold.}
    \label{tab:gen_res_average_df}
\end{table*}

\begin{table*}[h!]
    \centering
    \small
    \begin{adjustbox}{max width=\textwidth}
    \begin{tabular}{@{}llcccc|cccc@{}}
    \toprule
                                        &                 & \multicolumn{4}{c|}{\textbf{Articles}}                                                                                                            & \multicolumn{4}{c}{\textbf{Chunks}}                                                                                                                                   \\ \midrule
                                        &                 & \multicolumn{1}{l}{\textbf{ROUGE-LSum}} & \multicolumn{1}{l}{\textbf{BARTScore}} & \multicolumn{1}{l}{\textbf{SummaC}} & \textbf{GoldSim} & \multicolumn{1}{l}{\textbf{ROUGE-LSum}} & \multicolumn{1}{l}{\textbf{BARTScore}} & \multicolumn{1}{l}{\textbf{SummaC}} & \multicolumn{1}{l}{\textbf{GoldSim}} \\
    \multirow{5}{*}{\begin{turn}{90}\textbf{zero-shot}\end{turn}} 
                                        & mistral-v0.1    & \textbf{0.19}    & -2.16  & \textbf{0.35}    & -3.02  & \textbf{0.19} & -2.20  & \textbf{0.35} & -3.05                                        \\
                                          & mistral-v0.2    & 0.13  & -2.55  & 0.32  & -3.12  & 0.15  & -2.50  & 0.33  & -3.11 \\
                                          & llama3-8b   & 0.05  & -3.21  & 0.30  & -3.54  & 0.04  & -3.37  & 0.32  & -3.65 \\
                                          & llama2-7b   & 0.16  & -2.12  & 0.32  & -2.84  & \textbf{0.19}  & -2.06  & 0.33  & -2.87  \\
                                          & llama2-13b  & 0.17  & \textbf{-1.89}  & 0.34  & \textbf{-2.73}   & \textbf{0.19}  & \textbf{-1.89} & \textbf{0.35} & \textbf{-2.77} \\ \midrule
    \multirow{5}{*}{\begin{turn}{90}\textbf{one-shot}\end{turn}}  
                                        & mistral-v0.1    & \textbf{0.16}    & -2.38  & \textbf{0.33}  & -3.02  & 0.17  & -2.47  & 0.32  & -3.07 \\
                                        & mistral-v0.2 & 0.12  & -2.61  & 0.31  & -3.16  & 0.13  & -2.63  & 0.36  & -3.16  \\
                                        & llama3-8b  & 0.13 & \textbf{-2.25} & \textbf{0.33}   & -2.98 & 0.14 & -2.26 & \textbf{0.33} & -2.97 \\
                                        & llama2-7b  & \textbf{0.16}   & -2.35 & 0.32 & -2.93 & \textbf{0.19}   & -2.34 & 0.31 & -2.89 \\
                                        & llama2-13b & \textbf{0.16}   & -2.31 & 0.32 & \textbf{-2.90} & 0.17 & \textbf{-2.22} & 0.31  & \textbf{-2.79}   \\ \midrule
    \begin{turn}{90}\textbf{ft}\end{turn} & llama2-13b & 0.10 & -2.08 & 0.39 & -2.59 & 0.10 & -2.21 & 0.38 & -2.69 \\ \bottomrule
    \end{tabular}
    \end{adjustbox}
    \caption{Generation results per LLM, \textit{averaged across the three datasets} (neutral, SMP, emotional). Retrieved articles or chunks were employed in the generation. The best results for each generation configuration are in bold.}
    \label{tab:average_llm_chunks}
\end{table*}

\paragraph{Generation with Silver KB}
\label{subsec:gen_extra_ev}
Finally, we tackled the %
scenario where the useful information %
is spread across several documents: this lifts the constraint of existing datasets wherein a claim is paired to a single article. 
We used all documents from the Silver KB (Section~\ref{par:knowledge_base}) and the LLM-Retriever/\texttt{Llama-2-13b} setup (best-performing in the above).
We focused solely on the chunk-based configuration as the information required to build a verdict is \emph{(i)} often distributed across multiple extra documents, %
and \emph{(ii)} it is more likely to be located in specific sections of these extra evidence articles. Results (Table~\ref{tab:extra-evidence-gen-res}) show that, 
except for ROUGE-LSum, \texttt{Llama-2-13b}'s performance is consistently slightly worse when compared to generation using the Gold KB.\footnote{Compare with Tables \ref{tab:average_llm_chunks}, \ref{tab:gen_res_average_df}, and Appendix~\ref{app:generation_results} - Table~\ref{tab:complete_gen_results}.}
Still, a qualitative analysis showed that using the Silver KB resulted in verdicts that, in most cases, were consistent with the claim, faithful to the context, and informative. 

The lower results can be explained by the substantial difference between the Gold and the Silver KBs: a gold fact-checking article refers to a single claim and contains all the information needed to generate the verdict. Therefore, when using the Gold KB, out of a total of ten chunks, a robust retriever  -- such as LLM-Retriever -- can identify a larger number of informative chunks. Conversely, in the case of the Silver KB, for each claim, on average there are four related articles (see Appendix~\ref{app:extra_evidence}, Table~\ref{tab:extra_stats}) that are most likely to provide partial information about the verdict. 
Therefore, realistic retrieval scenarios for professional fact-checkers involve large, informationally sparse, and repetitive document collections, meaning 10 retrieved chunks may lack sufficient information for a good verdict.

\begin{table}[]
    \centering
    \begin{adjustbox}{max width=\columnwidth}
    \begin{tabular}{clcccc}
    \toprule
    \multicolumn{1}{l}{}         &                              & \textbf{ROUGE-LSum}            & \textbf{BARTScore}             & \textbf{SummaC}               & \textbf{GoldSim}      \\ \midrule
                                 & Neutral                     & 0.23                         & -2.17                         & 0.28                         & -3.04                         \\
                                 & SMP                          & 0.21                         & -2.30                         & 0.28                         & -2.84                         \\
                                 & Emotional                    & 0.19                         & -2.50                         & 0.26                         & -2.73                         \\  
    \multirow{-4}{*}{\begin{turn}{90}\textbf{zero-shot}\end{turn}}  
                                 & \cellcolor[HTML]{D9D9D9}Mean & \cellcolor[HTML]{D9D9D9}0.21 & \cellcolor[HTML]{D9D9D9}-2.32 & \cellcolor[HTML]{D9D9D9}0.28 & \cellcolor[HTML]{D9D9D9}-2.87 \\ \cmidrule(l){2-6}
                                 & Neutral                     & 0.24                         & -2.39                         & 0.24                         & -3.02                         \\
                                 & SMP                          & 0.21                         & -2.57                         & 0.25                         & -2.90                         \\
                                 & Emotional                    & 0.18                         & -2.68                         & 0.26                         & -2.76                         \\  
    \multirow{-4}{*}{\begin{turn}{90}\textbf{one-shot}\end{turn}}   
                                 & \cellcolor[HTML]{D9D9D9}Mean & \cellcolor[HTML]{D9D9D9}0.21 & \cellcolor[HTML]{D9D9D9}-2.55 & \cellcolor[HTML]{D9D9D9}0.25 & \cellcolor[HTML]{D9D9D9}-2.89 \\ \cmidrule(l){2-6}
                                 & Neutral                     & 0.12                         & -2.18                         & 0.39                         & -3.00                         \\
                                 & SMP                          & 0.13                         & -2.64                         & 0.26                         & -2.59                         \\
                                 & Emotional                    & 0.13                         & -2.97                         & 0.25                         & -2.70                         \\
    \multirow{-4}{*}{\begin{turn}{90}\textbf{fine tuned}\end{turn}} 
                                 & \cellcolor[HTML]{D9D9D9}Mean & \cellcolor[HTML]{D9D9D9}0.13 & \cellcolor[HTML]{D9D9D9}-2.60 & \cellcolor[HTML]{D9D9D9}0.30 & \cellcolor[HTML]{D9D9D9}-2.77 \\ \bottomrule
    \end{tabular}
    \end{adjustbox}
    \caption{Generation results with Silver KB.}
    \label{tab:extra-evidence-gen-res}
\end{table}

\paragraph{Human Evaluation}

Automatic metrics for NLG evaluation are known to correlate poorly with human judgments. Several works showed how optimizing for such metrics (e.g., ROUGE) is largely suboptimal \cite{paulus2018a, scialom2019answers}, and they suffer from weak interpretability and failure to capture nuances \cite{10.1145/3485766}.
Therefore, we also provide a comprehensive human evaluation of the generated verdicts.

\begin{figure}
    \centering
    \includegraphics[width=\columnwidth]{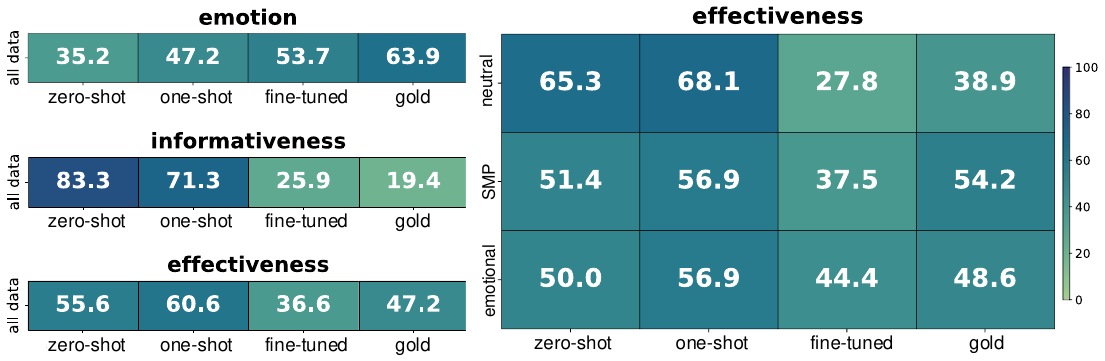}
    \caption{Human evaluation results: Percentages of preference for the four generation setups across the datasets.}
    \label{fig:human_evaluation_results}
\end{figure}

We enlisted three expert evaluators,\footnote{A senior researcher and two MSc graduates; all volunteer evaluators are proficient in English, experts in NLP, and knowledgeable about social media platforms' communication styles and dynamics, particularly in the context of misinformation.} and we focused on the data generated using Llama2-13b and the LLM-Retrieval over the Gold KB, as this configuration yielded the best results in both the retrieval and generation phases (see Sections~\ref{par:retr_extra_ev},~\ref{subsec:gen_extra_ev}). The evaluators were provided with pairs of verdicts (either gold or generated using zero-shot, one-shot, or fine-tuned models) and their corresponding claims. They were instructed to assess the best verdict based on three aspects: effectiveness, informativeness, and emotional/empathetic coverage. We sampled 72 claims equally distributed among the test sets, and provided the evaluators with six combinations of verdict pairs, compounding to 432 samples to be evaluated on the three evaluation dimensions; thus, the reported human evaluation is based on a total of 1296 data points.
In Figure~\ref{fig:human_evaluation_results} (left) we report how many times, in percentage, the human annotators preferred each of the four verdict generation setups (gold, zero-shot, one-shot, fine-tuning) across the three claim styles (neutral, SMP, emotional). 
The interannotator agreement (Cohen’s $\kappa$) was 0.7, indicating good agreement.

Results show that zero-shot and one-shot approaches were largely preferred in terms of informativeness. Gold and fine-tuned configurations were considered the best in terms of emotional matching between the claim and the related verdicts, which can be explained by the fact that fine-tuned models learned to mimic the emotional style of the gold training data. This is supported by the effectiveness evaluation in Figure~\ref{fig:human_evaluation_results} (right), where the delta between the preferences for zero/one-shot configurations and gold/fine-tuned ones is higher for neutral claims and it decreases as the emotional component increases. Nevertheless, on average, zero and one-shot configurations remain the most preferred. 
This result could stem from FullFact's verdicts' intended use alongside articles, not as standalone social corrections, and consequently aren't completely self-contained in grounding, with the full context available on the article page.
\footnote{Further details are provided in Appendix~\ref{app:human_evaluation_details}.} A fine-grained analysis of annotator agreement across the three aspects revealed a high $\kappa$ score for informativeness (0.8), and moderately strong scores for the more subjective dimensions of emotional coverage and effectiveness (both at 0.6).

\section{Conclusions}
The fight against misinformation has become increasingly demanding, amplified by the vast reach of popular social media.
With this work, we provided a comprehensive overview of how RAG-based approaches can effectively be employed for the automatic generation of verdicts when addressing realistic fact-checking scenarios. %
We thus present an extensive experimental analysis that explores different approaches at each stage of these pipelines, accounting for diverse computational constraints relevant to researchers, practitioners, and fact-checking organizations.

Our results show that LLM-based retrievers consistently demonstrate superior adaptability and performance in retrieval tasks, outperforming other methods. However, they face challenges when dealing with heterogeneous knowledge bases, where performance declines. Hybrid retrieval approaches offer a cost-effective alternative, while dense retrievers remain robust to stylistic variations. Notably, incorporating fact extraction modules enhances retrieval effectiveness across setups; thus, query preprocessing is particularly beneficial for addressing more complex claims. Although claim complexity reduces generation accuracy, larger models achieve greater faithfulness to context and alignment with gold-standard verdicts. Human evaluations also favour zero-shot and one-shot approaches, particularly for their informativeness.

\section{Limitations}
The data employed are limited to the English language only. However, we believe that the proposed RAG-based pipeline can still be adapted to a multilingual setting by adopting appropriate/multilingual retrievers and LLMs. 

Moreover, the datasets employed in the study are closely associated with FullFact topic domain (and style) and are confined to the SMP/emotional communication style adopted in VerMouth dataset. However, to the best of our knowledge, FullFact and VerMouth are the sole available aligned datasets that allow a thorough examination of the impact of the claim's communication style on a RAG-based verdict production task. Nonetheless, we are confident that the single modules employed in our work can be easily adapted to novel domains or SMP communication styles.

Prompt engineering in this study followed an iterative process, adapting existing prompts from prior RAG-based applications to fit the specific requirements of our task—for instance, aligning with the emotional framing of claims as proposed by Russo et al. (2023). However, we did not perform an exhaustive exploration of alternative prompt designs. This decision was driven by the absence of established metrics for evaluating prompt quality in this context and the high cost associated with conducting extensive human evaluations. Importantly, our primary objective was not to achieve state-of-the-art performance in RAG-based verdict generation, but rather to surface key challenges and identify effective strategies across a range of realistic configurations.

In this work, we focus on utilizing the extra evidence provided by fact-checking articles as a proxy for using the web as a knowledge base. This approach allows us to bypass the inherently complex challenges of determining source relevance and reliability, issues beyond the scope of this paper. Consequently, we strike a balance between leveraging external information and maintaining manageable complexity. Furthermore, our experiments reveal that even this constrained setting presents significant challenges. Therefore, we argue that mastering the use of extra evidence in this context is a crucial first step before moving towards more sophisticated and resource-intensive methods.

\section{Ethical Statement}
Our work is motivated by the potential to improve the accuracy and efficiency of automated fact-checking systems. However, we acknowledge that the development of such technologies can potentially, as any human artefact, be exploited by malicious actors.
In this case, the technological building blocks (e.g., the LLMs) can be tuned to accomplish goals opposite to ours (e.g., generate persuasive fake news). We argue that, while malicious actors would keep pursuing their goals regardless of the community efforts, our work provides a contribution to keep up with their pace and foster advancements by relying exclusively on publicly available data and models, and by publicly releasing novel artefacts (e.g., the fine-tuned \texttt{Llama2-13b} checkpoints).

\section*{Acknowledgments}
This work was partly supported by: the AI4TRUST project - AI-based-technologies for trustworthy solutions against disinformation (ID: 101070190), the European Union’s CERV fund under grant agreement No. 101143249 (HATEDEMICS), the European Union’s Horizon Europe research and innovation programme under grant agreement No. 101135437 (AI-CODE).

\bibliography{final_custom}
\bibliographystyle{acl_natbib}
\newpage
\clearpage
\appendix

\section{Dataset Details}
\label{app:dataset_details}
In this work, we employed data from the FullFact \citep{russo-etal-2023-benchmarking} and VerMouth \citep{russo-etal-2023-countering} datasets. The FullFact dataset consists of \texttt{claim-article-verdict} triplets extracted from the FullFact website\footnote{\href{https://fullfact.org}{https://fullfact.org}}. The VerMouth dataset extends the FullFact dataset. Starting from FullFact's triplets, \citet{russo-etal-2023-countering} leveraged an \textit{author-reviewer pipeline} \citep{tekiroglu-etal-2020-generating} to rewrite the data according to social media platform style, either in a general manner or with an embedded emotional component. For more comprehensive details on the datasets, we encourage readers to refer to the original papers \citep{russo-etal-2023-benchmarking, russo-etal-2023-countering}.
In Table \ref{tab:data_distribution}, we detail the distribution of entries across the training, evaluation, and test sets for each dataset, namely FullFact \citep{russo-etal-2023-benchmarking} and VerMouth \citep{russo-etal-2023-countering}.

\begin{table}[h!]
\centering
\small
\begin{tabular}{@{}lllccc@{}}
\toprule
 &  &  & Train & Eval & Test \\ \midrule
\multicolumn{3}{c}{\textbf{FullFact}} & 1470 & 184 & 174 \\ \midrule
\multirow{7}{*}{\begin{turn}{90}\textbf{VerMouth}\end{turn}} & \multicolumn{2}{c}{\textbf{SMP}} & 1470 & 184 & 174 \\ \cmidrule(l){2-6} 
 & \multirow{6}{*}{\begin{turn}{90}\textbf{emotions}\end{turn}} & anger & 1265 & 158 & 158 \\
 &  & disgust & 1339 & 164 & 163 \\
 &  & fear & 1440 & 179 & 171 \\
 &  & happiness & 1200 & 165 & 149 \\
 &  & sadness & 1404 & 173 & 171 \\
 &  & surprise & 1433 & 181 & 170 \\ \bottomrule
\end{tabular}
\caption{FullFact and VerMouth data distribution.}
\label{tab:data_distribution}
\end{table}

\section{Fact Extraction Module}
\label{app:fact_extraction}
In order to remove noise from VerMouth's claims \citep{russo-etal-2023-countering}, we prepend a fact extraction module before passing the claim to the retriever. To this end, we prompted \texttt{Llama-2-13b} and provided an example of the expected output (one-shot). Hereafter, we report the prompt employed:

\begin{mdframed}[roundcorner=8pt, linewidth=0pt, backgroundcolor=gray!20, innermargin=0pt, outermargin=0pt, skipabove=5pt, skipbelow=0pt, leftmargin=0pt, rightmargin=0pt, font=\small]
\texttt{\textbf{SYSTEM:}} Extract from the following text the main fact. Remove possible opinions or emotional statements. \\
Report results in the following format: FACT:[main fact] \\

\noindent Here there is an example: \\
\noindent TEXT: "I just heard about the Covid-19 vaccines \& sadly they don't seem to be very effective in preventing the virus. Really disappointing! \#vaccineineffective \#covid19vaccin " \\
\noindent FACT: "The Covid-19 vaccines offer very little protection against the disease." \\

\noindent \texttt{\textbf{USER:}} Now extract the main fact from the following text:
TEXT:\{claim\}
\end{mdframed}

To evaluate the performance of the fact extraction module, we randomly selected 70 instances, evenly distributed across the different claim types. An expert evaluator was provided with a list of claims from the VerMouth dataset along with the corresponding extracted facts generated by the \texttt{Llama-2-13b} model. The evaluator was then asked to assess whether the model had successfully identified and extracted the underlying fact. Results show that in only 3 cases (4\%), the model failed to extract the fact. Notably, in these instances, the original claims framed the information as an opinion rather than an objective fact. Consequently, the model reproduced the speaker’s opinion instead of isolating the factual content. An example is presented below.

\begin{quote}
    \textit{"As a student, the thought of having more teachers than necessary disgusts me. It's not about quantity, it's about quality education. Let's invest in our teachers and give them the support they need to make a real difference in students' lives. \#educationreform \#qualityoverquantity"}
    \end{quote}
    \begin{quote}
    \textit{"The speaker believes that investing in teachers and providing them with support is important for quality education."}
\end{quote}

\section{Extra Evidence Extraction Details}
\label{app:extra_evidence}
In Table~\ref{tab:extra_stats}, we present detailed information about the additional evidence extracted from FullFact fact-checking articles, used to approximate the realistic scenario in which a gold fact-checking article is not available or does not exist (yet). From the original FullFact fact-checking articles we removed links to social networks and the source URL of the claim. Indeed, the claims fact-checked by FullFact vary in nature, often originating from social media posts, images, videos, and sometimes misleading headlines. Consequently, the source of the claim might not always provide additional information beyond the claim itself that can be used for verification. Furthermore, even if the claim’s source contains extra text, the information can potentially be misleading. Therefore, following our ``reliability requirement'' we filtered out the claim sources.

\begin{table}[ht!]
\centering
\small
\begin{tabular}{@{}lccccc@{}}
\toprule
 & \textbf{extra art} & \textbf{extra} & \textbf{words} & \textbf{sent} & \textbf{chunks} \\ \midrule
\textbf{all} & 4093 & 4 & 970 & 38 & 69412 \\
\textbf{test} & 672 & 4 & 868 & 35 & 9983 \\ \bottomrule
\end{tabular}
\caption{Statistics for all additional evidence extracted from FullFact fact-checking articles and the test set used in our experiments. We report the total number of extra evidence documents (\textit{extra art}); the average number of extra documents per fact-checking article (\textit{extra}); the average number of words (\textit{words}) and sentences (\textit{sent}); and the total number of chunks (\textit{chunks}).}
\label{tab:extra_stats}
\end{table}

\section{Retrieval Experiments Details}

\subsection{Retrievers Details}
\label{app:retrievers-details}

For the LLM-Retrieval configuration, we employed \texttt{e5-mistral-7b-instruct}, making slight modifications to the original prompt to better align it with our task requirements. The following prompt was employed: 

\begin{mdframed}[roundcorner=8pt, linewidth=0pt, backgroundcolor=gray!20, innermargin=0pt, outermargin=0pt, skipabove=5pt, skipbelow=0pt, leftmargin=0pt, rightmargin=0pt, font=\small]
Instruct: Retrieve relevant documents to support or refute the given claim. \\
\noindent Query: "\{query\_str\}"
\end{mdframed}

\subsection{Retrieval Results}
\label{app:retrieval_results}
\begin{figure*}[htb!]
      \centering
      \includegraphics[width=\textwidth]{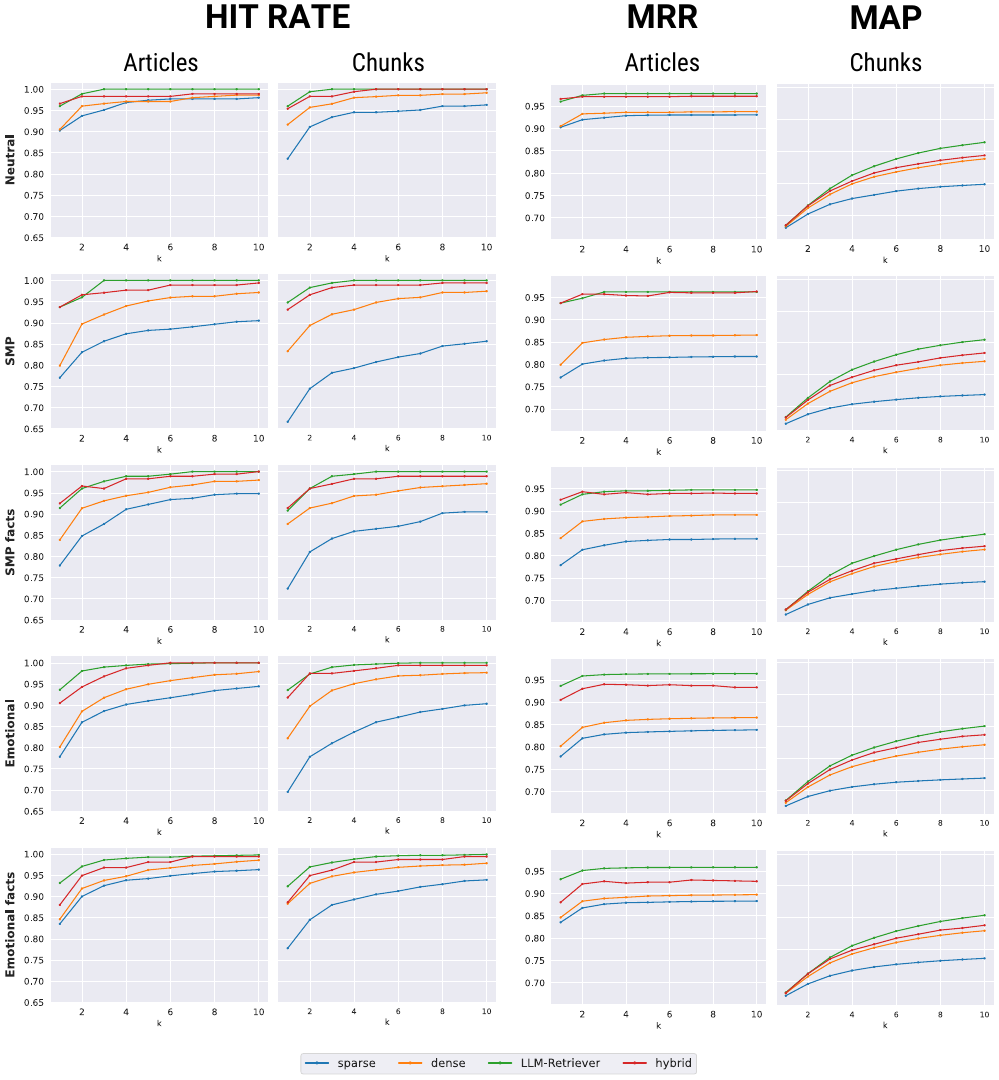}
      \caption{Retrieval results for each type of retriever (sparse, dense, LLM, hybrid) across Gold\_KB$_{art}$ and Gold\_KB$_{chunks}$ are presented for all claim styles, both with (SMP Facts, Emotional Facts) and without (neutral, SMP, emotional) claim pre-processing. The metrics reported include hit\_rate and MRR for retrieval over Gold\_KB$_{art}$, and hit\_rate and MAP for Gold\_KB$_{chunks}$, for increasing values of retrieved documents/chunks ($k=1,...,10$).}
      \label{fig:retrieval_results}
    \end{figure*}

In Figure~\ref{fig:retrieval_results}, we report hit\_rate, Mean Reciprocal Rank (MRR), and Mean Average Precision (MAP) for the retrieval experiments.

\section{Generation Experiments Details}
\label{app:generation_pahse}

\subsection{Model's instruction}
\label{app:model_instruction}
Hereafter, we report the instruction employed for the zero-shot setting. A similar instruction was modified by adding an example from the training sets in the one-shot configuration.

\begin{mdframed}[roundcorner=8pt, linewidth=0pt, backgroundcolor=gray!20, innermargin=0pt, outermargin=0pt, skipabove=5pt, skipbelow=0pt, leftmargin=0pt, rightmargin=0pt, font=\small]
\texttt{\textbf{SYSTEM:}}
Based on the provided context, respond to the claim, ensuring a thorough explanation. Use only the given context. Reply in no more than three sentences. Avoid mentioning the context in the reply. Match the communication style of the claim and address the possible emotional component present in it, if needed. If the context is insufficient, state that you don't know. Format your response as follows: \\ \\
\noindent Reply: [your\_reply] \\ \\
\texttt{\textbf{USER:}} The context information is provided below (in between xml tags). \\ \\
\noindent <context> \\ 
\noindent \{context\_str\} \\
\noindent </context> \\ \\
\noindent Claim: "\{query\_str\}"
\end{mdframed}

\subsection{Training Set Creation}
\label{app:training_set_creation}

In order to fine-tune the LLM for the RAG-based verdict production task, three main elements are needed: a claim, a gold answer, and the context comprising the knowledge needed to reply. In FullFact and VerMouth, the knowledge is present in the form of a fact-checking article. This comes in useful when the entire article is used as a context, but when working with chunks a proper selection of the most informative chunks must be performed. To this end, we started from the gold verdicts present in the two aforementioned datasets, and we ranked each article's chunks given the verdict information using a cross-encoder reranking model, i.e. the \texttt{BAAI/bge-reranker-large}\footnote{\href{https://hf.co/BAAI/bge-reranker-large}{https://hf.co/BAAI/bge-reranker-large}}.
Both for the articles and the chunk configurations, we add to the context of each training entry some negative examples, as in testing time the retrieved content might comprise articles or chunks that are not gold. To do so, we employed BM25 for retrieving 10 articles/chunks for each gold verdict and selected the non-gold retrieved context. An example of training input is provided in Table~\ref{tab:train_data_example}.

\begin{table*}[b!]
    \small
    \centering
    \resizebox{\textwidth}{!}{%
    \begin{tikzpicture}
    \node (table) [inner sep=0pt] {
    \begin{tabular}{p{\textwidth}}
    \\  \texttt{<s>[INST] <<SYS>>}
Based on the provided context, respond to the claim, ensuring a thorough explanation. Use only the given context. Reply in no more than three sentences. Avoid mentioning the context in the reply. Match the communication style of the claim and address the possible emotional component present in it, if needed. If the context is insufficient, state that you don't know. Format your response as follows:
\\\\
Reply: [your\_reply]
\\\\
\texttt{<</SYS>>}
\\\\
The context information is provided below (in between xml tags).
\\\\
\textbf{<context>}
\\\\
\ctext[RGB]{162,228,184}{A meme shared on Facebook features actor John Krasinski in The Office with a whiteboard with edited text, which says: “3 countries refused the covid vaccine”, followed by: “Now all 3 of their presidents have died unexpectedly”. Beneath the image are the names of the former presidents of Haiti (Jovenel Moïse), Tanzania (John Magufuli) and Zambia (Kenneth Kaunda).}
\\\\
\ctext[RGB]{162,228,184}{The president did not refuse the Covid-19 vaccines for Zambia. In fact, in March 2021, the Zambian health minister announced plans to vaccinate all over 18s in the country. Similar claims have been fact checked before.}
\\\\
\ctext[RGB]{255,114,118}{This survey covers households in England and Wales and so does not cover groups (such as those living in student halls of residence), who have “potentially high proportions of drug use”, meaning the true figure could be higher. Comparing England \& Wales to other countries in Europe is difficult because not all countries have up to date data.}
\\\\
\ctext[RGB]{255,114,118}{It’s correct that cocaine use among 16 to 24 years olds in England and Wales is at its highest level for around a decade. In 2017/18 6\% said they had used at least once in the previous year. The claim referred to Britain, but used data covering only England \& Wales. We’re focusing on England \& Wales as data for Scotland and Wales are not available for the most recent year.}
\\\\
\ctext[RGB]{162,228,184}{There is no evidence to suggest that the death was related to Mr Magufuli’s stance on the Covid-19 vaccines. There has been some speculation from Tanzanian opposition leaders, and on social media, that Mr Magufuli’s death may have been caused by Covid-19, however this has been discredited. President Kaunda died of pneumonia at a military hospital in Lusaka in June 2021, age 97.}
\\\\
\ctext[RGB]{162,228,184}{President Magufuli reportedly said that home treatments such as steam inhalation were preferable to “dangerous foreign vaccines”, and in February 2021 the country’s health minister said that Tanzania had no plans to accept Covid-19 vaccines.  Mr Magufuli’s successor,  president Samia Suluhu Hassan announced that the president’s death in March 2021 was due to heart disease.}
\\\\
\ctext[RGB]{162,228,184}{Arrests have been made but there are still many unknown details about the assasination. There is no evidence to suggest that there is a link to the lack of progress made regarding Haiti’s vaccine roll out.}
\\\\
\ctext[RGB]{162,228,184}{President Moïse did not explicitly refuse all of the Covid-19 vaccines, but the country did initially refuse the AstraZeneca vaccine due to safety concerns.  There have also been administrative and supply difficulties around the arrival of vaccinations to Haiti, and concerns about the logistics of organising the vaccine roll out. Mr Moïse was fatally shot at home on 7 July 2021.}
\\\\
\ctext[RGB]{162,228,184}{The posts suggest that the death of these three men is linked to a refusal of the Covid-19 vaccines. There is no evidence to support this claim. Haiti is eligible for provision of Covid-19 vaccines through the Covax programme, but roll out has been slow, with Haiti yet to give any doses to residents, and only receiving its first vaccines this week.}
\\\\
\ctext[RGB]{255,114,118}{The Royal College of Obstetricians \& Gynaecologists (RCOG) says that there is “no evidence” to suggest that the Covid-19 vaccines affect fertility. Similar statements have been issued by the British Fertility Association, the NHS and Public Health England. The RCOG also states that there is no biologically plausible mechanism by which the vaccine would cause fertility problems.}
\\\\
\textbf{</context>}
\\\\
\textbf{Claim:} "The presidents of Haiti, Tanzania, \& Zambia all passed away suddenly \& unexpectedly. Reports suggest it may be connected to their refusal to accept the Covid-19 vaccines for their countries. This is a tragic loss for the world \& for the people of those countries. \#vaccinesaveslives \#RIP" \texttt{[/INST]}
\\\\
\textbf{Reply:} "I'm sorry to hear about the recent passing of the presidents of Haiti, Tanzania, and Zambia. However, it's important to clarify that the president of Zambia did not refuse vaccines and passed away from pneumonia at a ripe old age, while the president of Tanzania passed away from a heart condition. The assassination of the president of Haiti is a tragic event, but there is no evidence linking it to vaccine refusal. Let's honor their legacies and focus on ways to prevent further loss of life."
\\
    \end{tabular}
    };
    \draw [rounded corners=.5em, very thick] (table.north west) rectangle (table.south east);
    \end{tikzpicture}}
    \caption{Training example for \texttt{Llama-2-13b} model. The positive passages are highlighted in green, while negative in red.}
    \label{tab:train_data_example}
    \end{table*}

\subsection{Fine-Tuning Details}
\label{app:fine_tuning_details}

We fine-tuned the \texttt{Llama-2-13b}\footnote{\href{https://hf.co/meta-llama/Llama-2-13b-chat-hf}{https://hf.co/meta-llama/Llama-2-13b-chat-hf}} chat model on different subsamples of training data from the FullFact and VerMouth datasets. From the training dataset, created following the procedure explained in Section \ref{app:training_set_creation}, we randomly extracted 200 entries each from the FullFact and SMP datasets. For the emotional datasets, we sampled 35 entries for each emotion, totalling 210 training entries. An example of input is shown in Table~\ref{tab:train_data_example}.

All models were trained on a single Ampere A40 with 48GB memory using the QLoRA strategy \citep{qlora}, with a low-rank approximation set to 64, a low-rank adaptation set to 16, and a dropout rate of 0.1. Evaluation steps were set at 25, and the batch size was 4. All models were trained for 3 epochs with a learning rate of 10$^{-4}$.

\subsection{Generation Results}
\label{app:generation_results}
In Table~\ref{tab:complete_gen_results} we report the complete results for zero-shot and one-shot experiments using chunks and articles as information context. 

\begin{table*}
\centering
\small
\begin{adjustbox}{max width=\textwidth}
\begin{tabular}{@{}llcccccccc@{}}
\toprule
                                          &                  & \multicolumn{4}{|c|}{\textbf{Articles}}                                                                                     & \multicolumn{4}{c}{\textbf{Chunks}}                                                                             \\ \midrule
\multicolumn{10}{c}{\textbf{Zero-shot}}                                                                                                                                                                                                                                                                     \\ \midrule
\textbf{model}                            & \textbf{dataset} & \textbf{ROUGE-LSum}       & \textbf{BARTScore}         & \textbf{SummaC}           & \multicolumn{1}{l|}{\textbf{GoldSim}} & \textbf{ROUGE-LSum}       & \textbf{BARTScore}         & \textbf{SummaC}           & \textbf{GoldSim}           \\ \midrule
\multirow{3}{*}{\textbf{mistral-7b-v0.1}} & fullfact         & 0,216                     & -1,948                     & 0,381                     & \multicolumn{1}{l|}{-3,101}           & 0,202                     & -1,992                     & 0,361                     & -3,112                     \\
                                          & SMP              & 0,185                     & -2,153                     & 0,333                     & \multicolumn{1}{l|}{-2,990}           & 0,190                     & -2,207                     & 0,351                     & -3,060                     \\
                                          & emotional        & 0,174                     & -2,389                     & 0,322                     & \multicolumn{1}{l|}{-2,953}           & 0,188                     & -2,411                     & 0,327                     & -2,990                     \\ \cmidrule(l){2-10}
\multirow{3}{*}{\textbf{mistral-7b-v0.2}} & fullfact         & 0,130                     & -2,315                     & 0,331                     & \multicolumn{1}{l|}{-3,084}           & 0,144                     & -2,251                     & 0,334                     & -3,101                     \\
                                          & SMP              & 0,140                     & -2,516                     & 0,327                     & \multicolumn{1}{l|}{-3,144}           & 0,145                     & -2,453                     & 0,335                     & -3,078                     \\
                                          & emotional        & 0,131                     & -2,803                     & 0,312                     & \multicolumn{1}{l|}{-3,134}           & 0,145                     & -2,789                     & 0,317                     & -3,143                     \\ \cmidrule(l){2-10}
\multirow{3}{*}{\textbf{llama3-8b}}       & fullfact         & 0,050                     & -3,143                     & 0,310                     & \multicolumn{1}{l|}{-3,516}           & 0,035                     & -3,439                     & 0,319                     & -3,648                     \\
                                          & SMP              & 0,047                     & -3,330                     & 0,295                     & \multicolumn{1}{l|}{-3,644}           & 0,032                     & -3,461                     & 0,328                     & -3,715                     \\
                                          & emotional        & 0,054                     & -3,142                     & 0,282                     & \multicolumn{1}{l|}{-3,466}           & 0,041                     & -3,217                     & 0,305                     & -3,575                     \\ \cmidrule(l){2-10}
\multirow{3}{*}{\textbf{llama2-7b}}       & fullfact         & 0,168                     & -1,979                     & 0,330                     & \multicolumn{1}{l|}{-2,914}           & 0,181                     & -1,988                     & 0,330                     & -2,932                     \\
                                          & SMP              & 0,160                     & -2,118                     & 0,321                     & \multicolumn{1}{l|}{-2,861}           & 0,195                     & -2,022                     & 0,330                     & -2,898                     \\
                                          & emotional        & 0,159                     & -2,253                     & 0,311                     & \multicolumn{1}{l|}{-2,755}           & 0,197                     & -2,171                     & 0,318                     & -2,781                     \\ \cmidrule(l){2-10}
\multirow{3}{*}{\textbf{llama2-13b}}      & fullfact         & 0,176                     & -1,714                     & 0,353                     & \multicolumn{1}{l|}{-2,787}           & 0,195                     & -1,718                     & 0,355                     & -2,811                     \\
                                          & SMP              & 0,169                     & -1,849                     & 0,331                     & \multicolumn{1}{l|}{-2,714}           & 0,186                     & -1,879                     & 0,352                     & -2,752                     \\
                                          & emotional        & 0,156                     & -2,118                     & 0,323                     & \multicolumn{1}{l|}{-2,691}           & 0,183                     & -2,058                     & 0,338                     & -2,754                     \\ \midrule
\multicolumn{10}{c}{\textbf{One-shot}}                                                                                                                                                                                                                                                                     \\ \midrule
\multirow{3}{*}{\textbf{mistral-7b-v0.1}} & fullfact         & \multicolumn{1}{c}{0,173} & \multicolumn{1}{c}{-2,186} & \multicolumn{1}{c}{0,336} & \multicolumn{1}{c|}{-3,073}           & \multicolumn{1}{c}{0,180} & \multicolumn{1}{c}{-2,224} & \multicolumn{1}{c}{0,341} & \multicolumn{1}{c}{-3,166} \\
                                          & SMP              & \multicolumn{1}{c}{0,160} & \multicolumn{1}{c}{-2,369} & \multicolumn{1}{c}{0,339} & \multicolumn{1}{c|}{-3,043}           & \multicolumn{1}{c}{0,161} & \multicolumn{1}{c}{-2,523} & \multicolumn{1}{c}{0,325} & \multicolumn{1}{c}{-3,018} \\
                                          & emotional        & \multicolumn{1}{c}{0,145} & \multicolumn{1}{c}{-2,586} & \multicolumn{1}{c}{0,307} & \multicolumn{1}{c|}{-2,947}           & \multicolumn{1}{c}{0,162} & \multicolumn{1}{c}{-2,664} & \multicolumn{1}{c}{0,302} & \multicolumn{1}{c}{-3,012} \\ \cmidrule(l){2-10}
\multirow{3}{*}{\textbf{mistral-7b-v0.2}} & fullfact         & \multicolumn{1}{c}{0,134} & \multicolumn{1}{c}{-2,347} & \multicolumn{1}{c}{0,321} & \multicolumn{1}{c|}{-3,177}           & \multicolumn{1}{c}{0,136} & \multicolumn{1}{c}{-2,393} & \multicolumn{1}{c}{0,320} & \multicolumn{1}{c}{-3,206} \\
                                          & SMP              & \multicolumn{1}{c}{0,123} & \multicolumn{1}{c}{-2,630} & \multicolumn{1}{c}{0,301} & \multicolumn{1}{c|}{-3,160}           & \multicolumn{1}{c}{0,130} & \multicolumn{1}{c}{-2,623} & \multicolumn{1}{c}{0,304} & \multicolumn{1}{c}{-3,117} \\
                                          & emotional        & \multicolumn{1}{c}{0,116} & \multicolumn{1}{c}{-2,865} & \multicolumn{1}{c}{0,318} & \multicolumn{1}{c|}{-3,138}           & \multicolumn{1}{c}{0,125} & \multicolumn{1}{c}{-2,871} & \multicolumn{1}{c}{0,321} & \multicolumn{1}{c}{-3,148} \\ \cmidrule(l){2-10}
\multirow{3}{*}{\textbf{llama3-8b}}       & fullfact         & \multicolumn{1}{c}{0,133} & \multicolumn{1}{c}{-1,960} & \multicolumn{1}{c}{0,351} & \multicolumn{1}{c|}{-3,003}           & \multicolumn{1}{c}{0,135} & \multicolumn{1}{c}{-1,905} & \multicolumn{1}{c}{0,344} & \multicolumn{1}{c}{-2,988} \\
                                          & SMP              & \multicolumn{1}{c}{0,138} & \multicolumn{1}{c}{-2,322} & \multicolumn{1}{c}{0,323} & \multicolumn{1}{c|}{-3,011}           & \multicolumn{1}{c}{0,137} & \multicolumn{1}{c}{-2,377} & \multicolumn{1}{c}{0,321} & \multicolumn{1}{c}{-3,013} \\
                                          & emotional        & \multicolumn{1}{c}{0,131} & \multicolumn{1}{c}{-2,469} & \multicolumn{1}{c}{0,328} & \multicolumn{1}{c|}{-2,927}           & \multicolumn{1}{c}{0,134} & \multicolumn{1}{c}{-2,485} & \multicolumn{1}{c}{0,334} & \multicolumn{1}{c}{-2,902} \\ \cmidrule(l){2-10}
\multirow{3}{*}{\textbf{llama2-7b}}       & fullfact         & \multicolumn{1}{c}{0,171} & \multicolumn{1}{c}{-2,091} & \multicolumn{1}{c}{0,327} & \multicolumn{1}{c|}{-2,977}           & \multicolumn{1}{c}{0,197} & \multicolumn{1}{c}{-2,016} & \multicolumn{1}{c}{0,325} & \multicolumn{1}{c}{-2,889} \\
                                          & SMP              & \multicolumn{1}{c}{0,158} & \multicolumn{1}{c}{-2,408} & \multicolumn{1}{c}{0,305} & \multicolumn{1}{c|}{-2,922}           & \multicolumn{1}{c}{0,181} & \multicolumn{1}{c}{-2,467} & \multicolumn{1}{c}{0,304} & \multicolumn{1}{c}{-2,911} \\
                                          & emotional        & \multicolumn{1}{c}{0,147} & \multicolumn{1}{c}{-2,561} & \multicolumn{1}{c}{0,317} & \multicolumn{1}{c|}{-2,875}           & \multicolumn{1}{c}{0,191} & \multicolumn{1}{c}{-2,549} & \multicolumn{1}{c}{0,306} & \multicolumn{1}{c}{-2,869} \\ \cmidrule(l){2-10}
\multirow{3}{*}{\textbf{llama2-13b}}      & fullfact         & \multicolumn{1}{c}{0,185} & \multicolumn{1}{c}{-2,057} & \multicolumn{1}{c}{0,322} & \multicolumn{1}{c|}{-2,942}           & \multicolumn{1}{c}{0,199} & \multicolumn{1}{c}{-1,992} & \multicolumn{1}{c}{0,309} & \multicolumn{1}{c}{-2,840} \\
                                          & SMP              & \multicolumn{1}{c}{0,154} & \multicolumn{1}{c}{-2,355} & \multicolumn{1}{c}{0,314} & \multicolumn{1}{c|}{-2,920}           & \multicolumn{1}{c}{0,172} & \multicolumn{1}{c}{-2,235} & \multicolumn{1}{c}{0,316} & \multicolumn{1}{c}{-2,803} \\
                                          & emotional        & \multicolumn{1}{c}{0,139} & \multicolumn{1}{c}{-2,523} & \multicolumn{1}{c}{0,314} & \multicolumn{1}{c|}{-2,839}           & \multicolumn{1}{c}{0,152} & \multicolumn{1}{c}{-2,441} & \multicolumn{1}{c}{0,303} & \multicolumn{1}{c}{-2,737} \\ \bottomrule
\end{tabular}
\end{adjustbox}
\caption{Complete results for each model tested on the three datasets in zero-shot and one-shot settings. Results for both chunks and article configurations are reported.}
\label{tab:complete_gen_results}
\end{table*}

\subsection{Generation Examples}
\label{app:examples}

In Table \ref{tab:fullfact_example} and \ref{tab:anger_example} we show examples of generations with claims from both FullFact and VerMouth (anger emotion) datasets. Each table comprises the following information: the claim; the gold verdict; the generations with \texttt{Llama-2-13b-chat} model in zero-shot, one-shot, and fine-tuning settings; the relevant evidence retrieved (either chunks, Table \ref{tab:fullfact_example}, or articles, Table \ref{tab:anger_example}).

\begin{table*}
\footnotesize
\centering
\begin{tabular}{p{0.30\linewidth}p{0.30\linewidth}p{0.30\linewidth}}
\toprule
\multicolumn{3}{p{0.90\linewidth}}{\textbf{CLAIM: }Research says that nearly half of people who catch Covid develop depression afterwards.} \\ \midrule
\multicolumn{3}{p{0.90\linewidth}}{\textbf{GOLD VERDICT: }This is not what the research says. It measured rates of depression among Covid survivors—but not whether they developed depression after catching the disease.} \\ \midrule
\multicolumn{1}{c}{\textbf{zero-shot}} & \multicolumn{1}{c}{\textbf{one-shot}} & \multicolumn{1}{c}{\textbf{fine-tuning}} \\ \midrule
The claim is partially accurate but misleading. The study in question found that 48\% of participants experienced moderate or severe depression, but it did not establish a causal link between Covid-19 and depression. The study also did not compare depression rates among those who had Covid with those who had not, nor did it consider the normal background rate of depression in Bangladesh. Additionally, the study did not find that Covid-19 directly causes depression. Instead, it suggests that people suffering from depression may be more likely to experience long-term symptoms after recovering from Covid-19. & The claim is misleading and not entirely accurate. The research in question found that nearly half of the participants reported experiencing depression "alongside" Covid, but it does not necessarily imply that Covid caused the depression. The study did not compare the prevalence of depression among those who caught Covid to those who did not, nor did it account for the normal background rate of depression in Bangladesh. Additionally, the study did not find any causal link between Covid and depression. The researchers themselves noted that the findings may not reflect the impact of Covid-19, especially for depression, which may have developed prior to the pandemic. It is important to be cautious when interpreting the results of scientific studies and not draw conclusions that go beyond the evidence. & The study found that 48\% of people who had Covid had symptoms of moderate or severe depression. But the researchers themselves say that this does not necessarily reflect the impact of Covid, and that it may have developed before the pandemic. \\ \midrule
\multicolumn{3}{p{0.90\linewidth}}{\begin{tabular}[c]{p{\linewidth}}\ctext[RGB]{162,228,184}{\textbf{1 - }The study in question explicitly said that its findings didn’t necessarily reflect the impact of Covid-19.The newspaper said: “Nearly half of people struck down with Covid-19 also develop depression, according to new research.”The research in question surveyed the mental health of 1,002 people in Bangladesh who had tested positive for Covid at some point in the past.}\\  \ctext[RGB]{162,228,184}{\textbf{2 - }These people were a self-selected sample, meaning they heard about the survey and wanted to take part, so they might not be representative of people as a whole.The study found that about 48\% had moderate or severe depression.However, it does not tell us that they “go on” to develop depression as a result of catching Covid, or even that they developed it after catching Covid.}\\  \ctext[RGB]{162,228,184}{\textbf{3 - }One of the study authors says it describes people suffering from depression “alongside” Covid.The researchers themselves say in the paper: “The findings may not reflect the impact of COVID-19, especially for depression, which may have developed prior to the COVID-19 pandemic.”There is evidence that rates of depression have increased in Great Britain and elsewhere during the pandemic.}\\  \ctext[RGB]{162,228,184}{\textbf{4 - }The Telegraph has muddled up the meaning of a scientific study on depression among people who have had Covid-19.}\\  \ctext[RGB]{162,228,184}{\textbf{5 - }The research doesn’t compare its findings with depression rates among these people before they got the disease, nor with people who hadn’t been infected at all, nor with the normal background rate of depression in Bangladesh (which may have risen during the pandemic).A press release from Anglia Ruskin University, which worked on the research, does not say that Covid was found to cause depression.}\\  \ctext[RGB]{162,228,184}{\textbf{6 - }There is other evidence that people may be at higher risk of developing mental health problems after recovering from Covid.The extent of this and why it might happen are not yet fully understood, however.}\\ \textbf{7 - } 26\% had symptoms of moderate or severe levels of depression with no real change across the first six weeks, which may be what the “one in four” claim was meant to refer to. Feelings of defeat and entrapment decreased over time, while positive wellbeing increased.Although the findings of the study were not as dramatic as claimed in the video, researchers said there was cause for concern.\\ \textbf{8 - } So Mr Hancock was right to say that one in 20 people would have long-term symptoms after a Covid infection—if you define “long-term” as more than eight weeks.However, he was wrong to compare this to the one-in-10 figure for people under 50, because this describes those who suffered from symptoms that lasted more than four weeks—rather than eight.As the research paper explains,\\ \textbf{9 - } Mr Hancock was also wrong to say that there was “no correlation” between the severity of the initial illness and the length of the symptoms afterwards.The research paper says people with long Covid were “more likely to have required hospital assessment in the acute period.” Overall, 13.9\% of the confirmed cases in the study had visited hospital. However, among those reporting symptoms for at least four weeks this rose to 31.5\%,\\ \textbf{10 - } and among those reporting symptoms for at least eight weeks it was 43.9\%.The app data also suggested that “Individuals reporting more than 5 symptoms in the first week (the median number reported) were significantly more likely to go on to experience LC28 symptoms lasting at least four weeks.”In short, more severe cases of Covid seemed more likely to result in long Covid afterwards.\end{tabular}} \\ \hline
\end{tabular}
\caption{Example of generation using FullFact claim, e5-mistral as a retriever and Llama-2-13b-chat for the generation of the verdict. In green are highlighted the gold chunks retrieved.}
\label{tab:fullfact_example}
\end{table*}

\begin{table*}
\footnotesize
\centering
\begin{tabular}{p{0.30\linewidth}p{0.30\linewidth}p{0.30\linewidth}}
\toprule
\multicolumn{3}{p{0.90\linewidth}}{\textbf{CLAIM: }I'm fuming over the fact that the worst violent and sexual criminals are let off the hook halfway through their sentence. How is this justice? I feel outraged and betrayed by the justice system." \#JusticeForVictims} \\ \midrule
\multicolumn{3}{p{0.90\linewidth}}{\textbf{GOLD VERDICT: }I understand your frustration and anger towards the justice system. It is important to note that judges in England and Wales can already choose to keep dangerous offenders in prison for up to two thirds of their sentence. I hope this information provides some clarity. \#JusticeForVictims} \\ \midrule
\multicolumn{1}{c}{\textbf{zero-shot}} & \multicolumn{1}{c}{\textbf{one-shot}} & \multicolumn{1}{c}{\textbf{fine-tuning}} \\ \midrule
It's understandable that you feel outraged and betrayed by the justice system, but it's important to note that the claim that the worst violent and sexual criminals are let off the hook halfway through their sentence is not entirely accurate. While it is true that some criminals are eligible for release halfway through their sentence, this is not automatic and is subject to the discretion of the judge. Additionally, judges already have the power to keep serious criminals in prison for longer than the normal halfway point, and they consider a variety of factors when determining the appropriate sentence. It's also worth noting that the proposed policy change by the Conservative party would not necessarily lead to longer sentences for all criminals, as judges may continue to sentence in the same way as they currently do. It's important to ensure that any changes to the justice system are fair, just, and proportionate, and that the needs of victims and the community are taken into account. & I understand your frustration, but it's important to note that the current sentencing procedure already allows for judges to keep serious criminals in prison for longer periods of time. The proposed change to mandatory two-thirds sentences would not necessarily lead to more time being served, as judges would still have discretion to sentence criminals based on the circumstances of the case. Additionally, the proposal only applies to the most serious violent and sexual offenders, and it's important to remember that the justice system is designed to balance punishment with rehabilitation and public protection. Let's focus on ensuring that victims receive the support and justice they deserve, while also ensuring that the system is fair and effective for all parties involved. & Thank you for sharing your concerns. It's important to note that the justice system is designed to provide a balance between punishment and rehabilitation. While it's understandable to feel outraged when it comes to serious crimes, it's important to remember that the current sentencing system already allows for judges to keep serious criminals in prison for longer periods of time if they deem it necessary. Additionally, the government has recently announced plans to make it mandatory for the most serious violent and sexual offenders to serve two-thirds of their sentence in prison, which may address your concerns. However, it's important to remember that the justice system is complex and there are many factor \\ \midrule
\multicolumn{3}{p{0.90\linewidth}}{\begin{tabular}[c]{p{\linewidth}}At the Conservative party conference last week, justice secretary Robert Buckland pledged to make it mandatory for the most serious violent and sexual offenders to serve two thirds of their sentence in prison (and not half the sentence as is normal). The Ministry of Justice told us this would apply to “the most serious sexual and violent offenders where the offence carries a maximum life sentence.” However, to suggest that the most serious criminals are automatically released from prison halfway through their sentences obscures the fact that judges already have discretion to keep serious criminals in prison for longer. There are various sentences a judge can hand down and in most cases sentences are non-custodial (where no time is served in prison). By far the most common sentence for crimes in England and Wales is a fine, but what’s of interest here are sentences which carry mandatory prison time. Typically in these cases a criminal will be given a standard determinate sentence. This usually requires them to spend half of their sentence in prison and the other half on license in the community, supervised by the probation service. For example, a standard two year sentence would involve one year in prison and one year on license. Being on licence means you can be recalled to prison if you breach the terms of your licence. As well as standard sentences, judges in England and Wales can hand down what are called ‘extended determinate sentences’ to criminals who commit any of over 100 serious offences. The judge can make this decision if: These offenders are either entitled to be released two thirds of the way through their sentence, or can apply for parole at that point. Life sentences work slightly differently. With a life sentence a criminal is required to spend a minimum time in prison and is then able to apply for parole. If they are released, they remain on license for the rest of their life. Compared to existing extended sentences, the Conservatives’ proposal appears to apply to criminals who commit a slightly different group of offences (those that carry a maximum of life rather than this list of serious offences). There is also apparently no requirement for a judge to determine if a criminal poses a risk to the public when giving this new kind of sentence. While the current sentencing procedure does not dramatically change the ability to put serious criminals in prison for two thirds  of their term, it would, in practice, significantly increase the number of criminals receiving two thirds sentences. That’s because judges rarely hand down extended sentences. The Ministry of Justice says that in 2018 there were around 4,000 standard sentences with halfway  release handed down to criminals who committed sexual or violent offences which carry the maximum penalty of life. By comparison, in 2018 judges in England and Wales handed down 398 extended sentences. There is an open question over whether the policy would in fact lead to serious criminals spending more time in prison, because it’s possible that judges could change how they currently sentence. [...]
\end{tabular}} \\ \hline
\end{tabular}
\caption{Example of generation using VerMouth anger claim, e5-mistral as a retriever and Llama-2-13b-chat for the generation of the verdict. The article has been cut for space reasons. The complete article's text can be found at \href{https://fullfact.org/crime/extended-sentences/}{https://fullfact.org/crime/extended-sentences/}}
\label{tab:anger_example}
\end{table*}

\section{Human Evaluation Details}
\label{app:human_evaluation_details}

\begin{figure*}
    \centering
    \includegraphics[width=\textwidth]{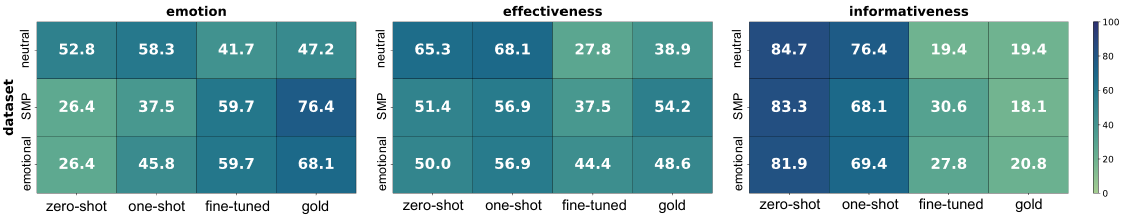}
    \caption{Complete results for the human evaluation. Each matrix refers to the results obtained for each verdict evaluation aspect. The matrices report how many times, in percentage, the human annotators preferred each of the four generation setups (gold, zero-shot, one-shot, fine-tuning).}
    \label{fig:human_eval_complete_matrix}
\end{figure*}

For the human evaluation of the generated verdict, we enrolled three volunteer evaluators. They were provided with pairs of verdicts (either gold or generated using zero-shot, one-shot, or fine-tuned models) and their corresponding claims. They were instructed to assess the best verdict based on three aspects: effectiveness, informativeness, and emotional/empathetic coverage. Hereafter, we list the tasks/questions that evaluators were required to follow when judging the verdict pair.

\begin{quote}
\vspace{5mm}
\centerline{\textbf{Informativeness}}
Tell which of the two verdicts contains more information supporting its stance.

\centerline{\textbf{Emotional Coverage}}
Some of the claims can express a variety of emotions. Tell which of the two verdicts better takes into consideration the emotion of the claim by responding with empathy.

\centerline{\textbf{Effectiveness}}
Which of the two is an overall better verdict (with respect to the claim) that could be used to answer the claim?

\end{quote}
Given that the claims presented may cover sensitive subjects, we have incorporated a cautionary note in the task description: \textit{"This task may contain text that some readers find offensive."}. Additionally, we briefed the evaluators on the study's objectives and assured them that all collected data would be anonymized and solely utilized for research purposes.

In Figure~\ref{fig:human_eval_complete_matrix} we report the full outcome of the human evaluation, showing the details of the selected verdicts over the three datasets according to emotional coverage, informativeness and effectiveness. 

\end{document}